\documentclass[letterpaper, 10pt, conference]{ieeeconf}
\IEEEoverridecommandlockouts 
\usepackage[utf8]{inputenc}
\usepackage{graphicx}

\usepackage{amsmath, amssymb}
\usepackage{balance}
\usepackage{hyperref}
\usepackage{multirow}
\usepackage{booktabs}
\usepackage{cite}
\usepackage{textgreek}
\usepackage{rotating} 
\usepackage[export]{adjustbox}
\usepackage{array}
\usepackage[caption=false,font=footnotesize]{subfig}
\DeclareGraphicsExtensions{.png, .jpg, .pdf}
\graphicspath{{images/}}
\newcommand{\bibpath}{./bib}
\title{Bridging the Ex-Vivo to In-Vivo Gap: Synthetic Priors \\for Monocular Depth Estimation in Specular Surgical Environments%
}

\begin{document}

\author{Ankan Aich$^{1}$, Emma D. Ryan$^{2}$,  Kris Moe$^{3}$, Isaac Schmale$^{2}$, Li-Xing Man$^{2}$, and Yangming Lee$^{1}$%
\thanks{$^{1}$RoCAL, Rochester Institute of Technology, Rochester, NY, USA, 14456}%
\thanks{$^{2}$Department of Otolaryngology Head and Neck Surgery, University of Rochester Medical Center, Rochester, NY, USA, 14618. }%
\thanks{$^{3}$University of Washington, Department of Otolaryngology–Head and Neck Surgery, Seattle, USA, 98195. }%
}

\maketitle
\thispagestyle{empty}
\pagestyle{empty}

\begin{abstract}
Accurate Monocular Depth Estimation (MDE) is critical for autonomous robotic surgery. However, existing self-supervised methods often exhibit a severe "ex-vivo to in-vivo gap": they achieve high accuracy on public datasets but struggle in actual clinical deployments. This disparity arises because the severe specular reflections and fluid-filled deformations inherent to real surgeries. Models trained on noisy real-world pseudo-labels consequently suffer from severe boundary collapse. To address this, we leverage the high-fidelity synthetic priors of the \textit{Depth Anything V2} architecture, which inherently capture precise geometric details, and efficiently adapt them to the medical domain using Dynamic Vector Low-Rank Adaptation (DV-LORA). Our contributions are two-fold. Technically, our approach establishes a new state-of-the-art on the public SCARED dataset; under a novel physically-stratified evaluation protocol, it reduces Squared Relative Error by over 17\% in high-specularity regimes compared to strong baselines. Furthermore, to provide a rigorous reality check for the field, we introduce \textbf{ROCAL-T 90} (Real Operative CT-Aligned Laparoscopic Trajectories 90), the first real-surgery validation dataset featuring 90 clinical endoscopic sequences with sub-millimeter ($< 1$mm) ground-truth trajectories. Evaluations on ROCAL-T 90 demonstrate our model's superior robustness in true clinical settings.
\end{abstract}

\begin{keywords}
Monocular Depth Estimation; Foundation Models; Synthetic-to-Real Adaptation; Autonomous Robotic Surgery
\end{keywords}

\section{Introduction}
\subsection{3D Perception in Robotic Surgery}
Precise 3D dense reconstruction is a fundamental requirement for the advancement of autonomous robotic surgery\cite{lin2021multi,TII18Random,JNSB18Completeness}. Beyond simple visualization, depth estimation serves as the geometric foundation for critical downstream tasks, including dynamic active constraints, intraoperative registration, soft tissue tracking, and augmented reality overlay \cite{RAL17GPR,mane2022single,IJCARS22Virtual}. In these safety-critical applications, the perception system must deliver robust geometric information from monocular endoscopic video in real-time, enabling the robotic agent to interact safely with the complex anatomy  \cite{tian2022kimera,florence2018dense,ICRA21SkillAssessment}.

\subsection{Challenges and the Ex-Vivo to In-Vivo Gap}
Despite its importance, robust depth estimation in Minimally Invasive Surgery (MIS) remains a formidable challenge due to the unique and hostile nature of the endoscopic environment. As comprehensively reviewed in \cite{Sensors24Recon}, surgical scenes violate nearly all photometric assumptions used in standard computer vision, and lead to failures of classical vision models \cite{Sensors24Recon,lee2026disentangling}. The environment is characterized by homogeneous tissue textures\cite{Neurocomputing13SLAMIDE}, rapid non-rigid deformations caused by respiration or instrument interaction \cite{ICRA18RNNPlan,lamarca2020defslam,IROS18RNNSoft,RAL17GPR}, and severe non-Lambertian effects—specifically, high-intensity specular reflections on wet surfaces \cite{mahmoud2016orbslam,okatani1997shape,JNSB17AnatomicalRegion}. 

While existing benchmarks like the SCARED dataset \cite{scared} have significantly driven progress in the field, they predominantly feature ex-vivo anatomy (e.g., porcine cadavers) that does not fully capture the chaotic reality of in-vivo human surgery. In actual clinical deployments, the continuous presence of active bleeding, irrigation, and dynamically pooling fluids creates extreme specular highlights and transparent surfaces that are largely absent from standard training sets \cite{JNSB18RelativeMotion}. Furthermore, the active structured-light sensors used to capture ground truth in these datasets frequently fail in regions of high specularity due to signal saturation \cite{Hao2021EndoDepth}. 

\begin{figure*}[t]
    \centering
    \includegraphics[width=0.6\linewidth]{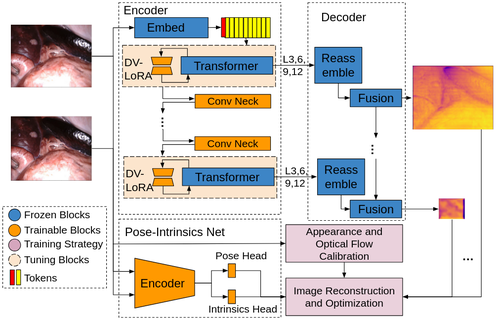}
\caption{Overview of the proposed self-supervised depth estimation framework. The \textbf{DepthNet} leverages a frozen \textit{Depth Anything V2} (DAv2) transformer backbone to preserve robust synthetic priors, effectively mitigating boundary collapse on thin surgical tools and transparent fluids. To bridge the domain gap, lightweight Dynamic Vector LoRA (DV-LORA) modules are injected into the attention layers to adapt to dynamic surgical illumination, while Convolutional Necks are interleaved to restore high-frequency tissue textures. Concurrently, a decoupled \textbf{Pose-Intrinsics Net} estimates 6-DoF camera motion and focal length, enabling self-supervised optimization via a view-synthesis objective on uncalibrated endoscopic video.}    \label{fig:architecture}
\end{figure*}

This discrepancy rises a critical "ex-vivo to in-vivo gap" in MDE research. Existing self-supervised models, evaluating themselves against these sanitized or incomplete datasets, often achieve high accuracy in simulated or ex-vivo environments but fail in true operative settings. Because the ground truth itself is missing in the hardest fluid-filled regions, models trained on noisy real-world pseudo-labels are inadvertently encouraged to hallucinate, leading to boundary collapse.

\subsection{Vision Backbones and the Need for Synthetic Priors}
In the broader computer vision community, the landscape of dense prediction has been revolutionized by Vision Transformers (ViTs) \cite{vaswani2017attention} and Foundation Models, such as the Segment Anything Model (SAM) \cite{kirillov2023segment} and the Depth Anything series \cite{Yang2024DepthAnythingV1}. These models, trained on massive-scale datasets, exhibit unprecedented zero-shot generalization. However, adapting these general-purpose breakthroughs to the surgical domain presents a significant bottleneck due to severe domain shift.

More critically, the pre-training paradigms of early foundation models inadvertently exacerbate this ex-vivo to in-vivo gap. Models like \textit{Depth Anything V1}, which forms the backbone of recent state-of-the-art surgical adapters like EndoDAC \cite{Cui2024EndoDAC}, rely heavily on massive real-world datasets and pseudo-labeling pipelines. Consequently, they inherit the very "label noise" that plagues real-world sensors \cite{Yang2024DepthAnythingV2}. The geometric priors learned from these noisy real-world labels consistently fail to resolve the fine-grained boundaries of thin surgical tools or reconstruct transparent, fluid-filled surfaces, leading to severe boundary collapse.

To break this reliance on flawed real-world supervision, we propose a framework that leverages the specific strengths of the \textit{Depth Anything V2} (DAv2) architecture \cite{Yang2024DepthAnythingV2}. Unlike its predecessors, DAv2 is pre-trained on high-fidelity \textit{synthetic} environments. We identify that these "synthetic priors" are uniquely suited to overcome the operative challenges outlined in Section I-B. Because synthetic datasets are generated via physical rendering engines, they provide mathematically precise depth supervision for complex optical properties—such as sharp tool edges, transparency, and specular reflections—that are inherently noisy or missing in real-world data. 

To bridge the synthetic-to-real domain gap, we efficiently adapt these synthetic priors to the medical domain using parameter-efficient Dynamic Vector Low-Rank Adaptation (DV-LORA) \cite{Cui2024EndoDAC}. Rather than training from scratch or blindly fine-tuning the entire network, this approach allows us to adapt to clinical textures and lighting while strictly preserving the model's robust, noise-free structural understanding.

\subsection{Contributions}
To rigorously address the ex-vivo to in-vivo gap and advance the state-of-the-art in endoscopic depth estimation, our main contributions encompass both technical advancements and a significant community benchmark:

\begin{enumerate}   
    \item \textbf{Synthetic Priors for Specular Environments:} We propose a parameter-efficient adaptation framework that successfully transfers the high-fidelity synthetic priors of the \textit{Depth Anything V2} foundation model to the medical domain. By integrating DV-LORA, we directly address the specific problem of boundary collapse on thin surgical tools and transparent fluids, maintaining precise geometric representations with a minimal parameter budget.
    
    \item \textbf{Physically-Stratified Evaluation Protocol:} Addressing the evaluation gap in existing literature \cite{Sensors24Recon}, we introduce a physically-stratified testing protocol on the public SCARED dataset \cite{scared}. By unsupervisedly clustering frames based on structured-light sensor failure rates, we rigorously quantify our model's robustness in high-specularity regimes, demonstrating a $>17\%$ error reduction compared to strong baselines trained on real-world priors.

    \item \textbf{The ROCAL-T 90 Clinical Benchmark:} We introduce ROCAL-T 90 (Real Operative CT-Aligned Localization and Trajectories 90), the first clinical validation dataset specifically designed to evaluate tracking in true operative settings. Derived from in-vivo endoscopic sinus surgeries, the dataset features 90 continuous video motion sequences meticulously aligned with patient-specific preoperative CT scans. This alignment establishes sub-millimeter ($< 1$mm) ground-truth trajectories, providing a rigorous reality check against the limitations of existing ex-vivo datasets. ROCAL-T 90 and its associated code will be publicly released upon final Institutional Review Board (IRB) approval.
\end{enumerate}

\section{Related Works}

\subsection{Self-Supervised Depth Estimation in Surgery}
Traditional self-supervised methods, pioneered by SfMLearner and Monodepth2 \cite{godard2019digging}, rely on photometric consistency to learn depth without ground truth. In the medical domain, numerous works have adapted this paradigm to endoscopy \cite{Hao2021EndoDepth}. Recently, Cui et al. introduced \textbf{EndoDAC} \cite{Cui2024EndoDAC}, a state-of-the-art framework that successfully adapted large-scale foundation models to the surgical domain using Parameter-Efficient Fine-Tuning (PEFT). Their work demonstrated that freezing a pre-trained backbone and injecting lightweight Dynamic Vector LoRA (DV-LORA) modules could achieve superior robustness compared to training from scratch. Our work builds directly upon this efficient adaptation architecture.

\subsection{Foundation Models: Real vs. Synthetic Priors}
The core differentiator of modern depth estimation is the pre-training data. The \textit{Depth Anything V1} backbone, utilized in the original EndoDAC, was trained using a semi-supervised pipeline on massive real-world datasets. While effective for general scenes, recent analyses suggest that V1's reliance on pseudo-labels introduces "label noise" around thin structures and transparent surfaces, as the teacher model often hallucinates in these regions \cite{Yang2024DepthAnythingV2}.

To address this, the recently introduced \textit{Depth Anything V2} (DAv2) shifts the training paradigm to high-fidelity \textbf{synthetic} environments \cite{Yang2024DepthAnythingV2}. This "synthetic prior" provides mathematically precise supervision for challenging optical properties (transparency, reflections) that are ubiquitous in surgery but noisy in real-world data. In this paper, we extend the EndoDAC framework by replacing the V1 backbone with DAv2, effectively combining Cui et al.'s efficient adaptation strategy with the superior geometric priors of synthetic pre-training.

\subsection{Surgical Depth and Tracking Datasets}
The development of surgical perception algorithms has historically been bottlenecked by a fundamental trade-off in data collection: the mutually exclusive pursuit of realistic surgical environments and high-precision ground truth \cite{Sensors24Recon}. Existing datasets generally fall into two categories. The first category prioritizes precise 6-DoF tracking and dense depth but captures them in simulated environments, such as silicone phantoms (e.g., C3VD \cite{bobrow2023colonoscopy}) or ex-vivo cadavers and tissues (e.g., SCARED \cite{scared}, EndoSLAM \cite{ozyoruk2021endoslam}). While immensely valuable, these environments fail to replicate the dynamic fluids, active bleeding, and erratic specular reflections of true in-vivo human procedures. The second category prioritizes clinical realism by capturing live in-vivo surgical videos (e.g., EndoMapper \cite{azagra2023endomapper}, Hamlyn \cite{Hamlyn}), but critically lacks absolute, sub-millimeter 6-DoF trajectory ground truth due to the extreme difficulty of integrating tracking hardware into the operating room without disrupting clinical workflows.

This persistent gap is largely driven by strict regulatory policies and patient safety constraints, which have historically prevented the capture of synchronized tracking data during live human surgeries\cite{Frontiers23DCL,Acta19ravenreview}. Through a multi-year collaboration with Stryker, we have successfully overcome these stringent technical and regulatory hurdles. This partnership enabled the fully IRB-compliant acquisition of synchronized endoscopic video and high-precision tracking data during live endoscopic sinus surgeries. To the best of our knowledge, our proposed ROCAL-T 90 dataset is the first to provide true in-vivo clinical sequences meticulously aligned with patient-specific preoperative CT scans, yielding sub-millimeter ground-truth trajectories. By breaking the simulation barrier, ROCAL-T 90 allows the community to evaluate algorithms against the definitive reality of the operating room.

\section{Methodology}
\subsection{Framework Overview}
Our framework builds upon the parameter-efficient adaptation strategy introduced in EndoDAC \cite{Cui2024EndoDAC}, but fundamentally redesigns the prior distribution to leverage the synthetic-to-real generalization capabilities of the \textit{Depth Anything V2} (DAv2) architecture. As illustrated in Figure \ref{fig:architecture}, the pipeline consists of two primary components: a \textbf{DepthNet} that predicts dense depth maps by injecting synthetic priors into the medical domain, and a decoupled \textbf{Pose-Intrinsics Net} that estimates the 6-DoF camera motion and focal length, enabling self-supervised training on uncalibrated clinical video.

\subsection{DepthNet: Injecting Synthetic Priors}
The core innovation of our DepthNet is the strategic preservation and adaptation of synthetic geometric priors to solve the severe boundary collapse and specularity artifacts prevalent in real-world endoscopic footage.

\subsubsection{Backbone Selection and Freezing}
Unlike previous architectures that initialize from real-world supervised models (e.g., DAv1), we utilize the DAv2 encoder, which is pre-trained on high-fidelity synthetic data. This initialization is critical for overcoming the ex-vivo to in-vivo gap in operative environments. In synthetic pre-training, complex optical phenomena—such as specular highlights on wet surfaces and transparent fluid pooling—are physically rendered alongside mathematically precise depth ground truth. Consequently, the network learns robust "synthetic priors" that inherently resolve thin geometric structures without hallucinating noise. To prevent catastrophic forgetting of these pristine geometric boundaries during adaptation to medical textures, we {freeze} the transformer backbone throughout the training process.

\subsubsection{Dynamic Vector Adaptation (DV-LORA)}
While the frozen backbone provides robust structural priors, it suffers from a severe domain shift when exposed to the specific photometric properties of in-vivo tissues (e.g., blood, homogeneous mucosa). To bridge this gap, we inject trainable Dynamic Vector LoRA (DV-LORA) modules into the attention layers. Endoscopic lighting is coaxial and highly dynamic, causing severe intensity fluctuations as the camera navigates tight anatomical cavities. Unlike standard LoRA, which relies on static low-rank matrices, DV-LORA introduces input-dependent dynamic vectors that scale the feature projection, allowing the model to adaptively recalibrate to extreme illumination changes. The updated weight matrix $\hat{W}$ is defined as:
\begin{equation}
\hat{W} = W_0 + \Lambda_v B \Lambda_u A,
\label{eq:dvlora}
\end{equation}
where $W_0 \in \mathbb{R}^{d \times k}$ is the frozen pre-trained weight, $A \in \mathbb{R}^{r \times k}$ and $B \in \mathbb{R}^{d \times r}$ are low-rank matrices ($r \ll d, k$), and $\Lambda_u, \Lambda_v \in \mathbb{R}^{r \times r}$ are diagonal matrices containing the learnable dynamic vectors. This configuration achieves superior synthetic-to-real transfer while adding only $\approx 1.6$M trainable parameters to the massive DAv2 backbone.

\subsubsection{High-Frequency Restoration}
Vision Transformers inherently act as low-pass filters, capturing global context but often over-smoothing local high-frequency details. In surgical scenes, recovering this high-frequency signal is paramount, as subtle tissue grain and micro-vascular structures are often the only visual cues available on otherwise homogeneous organ surfaces. Following \cite{Cui2024EndoDAC}, we interleave \textbf{Convolutional Neck} blocks after the 3rd, 6th, 9th, and 12th transformer layers. These blocks re-introduce local textural gradients into the feature stream before spatial reconstruction in the multi-scale decoder, effectively preserving critical anatomical landmarks.

\subsection{Self-Supervised Optimization}
Given the scarcity of dense depth ground truth in operative settings (as addressed by our ROCAL-T 90 benchmark), the network is optimized via a self-supervised view-synthesis objective.

\subsubsection{Decoupled Pose and Intrinsics}
Standard self-supervised SfM methods assume a fixed, pre-calibrated camera. However, clinical endoscopes frequently undergo zooming and focus adjustments intraoperatively, altering the focal length. We employ a decoupled Pose-Intrinsics network \cite{Cui2024EndoDAC}, which utilizes rotational constraints to disentangle focal length estimation from spatial translation, mitigating the scale-ambiguity degeneracy common in joint estimations.

\subsubsection{Loss Function}
The network minimizes the photometric reconstruction error between a target frame $I_t$ and a source frame $I_s$ warped into the target view ($I_{s \to t}$). We utilize the standard combination of Structural Similarity (SSIM) and L1 difference:
\begin{equation}
L_p = \alpha \frac{1 - \text{SSIM}(I_t, I_{s \to t})}{2} + (1 - \alpha) |I_t - I_{s \to t}|.
\end{equation}
To further regularize predictions in textureless tissue regions, an edge-aware smoothness loss $L_e$ \cite{godard2019digging} is applied. The final objective is the weighted sum: $L_{total} = L_p + \lambda L_e$.

\section{Experiments}
\subsection{Datasets and Evaluation Protocols}
To comprehensively evaluate our framework and expose the ex-vivo to in-vivo gap, we conduct experiments on both a standard public benchmark(SCARED) and our newly proposed clinical dataset.
\subsubsection{SCARED and Physically-Stratified Protocol}
We utilize the SCARED dataset \cite{scared}, a standard benchmark featuring ex-vivo porcine anatomy. Standard evaluation metrics often average performance across all pixels, masking catastrophic failures in critical regions. As noted in \cite{Sensors24Recon}, active sensors frequently fail in areas of high specular reflection, resulting in "invalid" pixels in the ground truth. To rigorously assess robustness against specularity, we introduce a \textbf{physically-stratified protocol}. We employ a Gaussian Mixture Model (GMM) to unsupervisedly cluster test frames based on the density of valid ground-truth pixels. This yields three regimes: \textbf{Hard} (High Specularity, $\approx 20\%$ valid), \textbf{Medium} , and \textbf{Easy} ($\approx 53\%$ valid).

\subsubsection{ROCAL-T 90 Clinical Benchmark}
To evaluate the true clinical viability of the models, we utilize our ROCAL-T 90 dataset. Unlike SCARED, this dataset consists of in-vivo endoscopic sinus surgery sequences. The ground-truth 6-DoF trajectories are acquired via a hardware tracking sensor rigorously aligned with patient-specific preoperative CT scans, providing sub-millimeter tracking accuracy in a chaotic, fluid-filled operative environment.

\subsection{Results on Public Benchmark (SCARED)}

\begin{table*}[!htbp]
\centering
\caption{Quantitative comparison of state-of-the-art methods on the SCARED dataset. 
}
\label{tab:sota_comparison}
\begin{tabular}{lccccc}
\toprule
\textbf{Method} & 
\textbf{Abs Rel} $\downarrow$ & 
\textbf{Sq Rel} $\downarrow$ & 
\textbf{RMSE} $\downarrow$ & 
\textbf{RMSE$_{\text{log}}$} $\downarrow$ & 
$\delta < 1.25$ $\uparrow$ \\
\midrule
SC-SfMLearner\cite{bian2019unsupervised} & 0.068 & 0.645 & 5.988 & 0.097 & 0.957 \\
Monodepth2\cite{monodepth2} & 0.069 & 0.577 & 5.546 & 0.094 & 0.948 \\
Fang \cite{fang2020towards} & 0.078 & 0.794 & 6.794 & 0.109 & 0.946 \\
Defeat-Net\cite{spencer2020defeat} & 0.077 & 0.792 & 6.688 & 0.108 & 0.941 \\
Endo-SfM\cite{ozyoruk2021endoslam} & 0.062 & 0.606 & 5.726 & 0.093 & 0.957 \\
AF-SfMLearner\cite{shao2022self} & 0.059 & 0.435 & 4.925 & 0.082 & 0.974 \\
Yang \cite{yang2024self} & 0.062 & 0.558 & 5.585 & 0.090 & 0.962 \\
DA (zero-shot)\cite{Yang2024DepthAnythingV1} & 0.084 & 0.847 & 6.711 & 0.110 & 0.930 \\
DA (fine-tuned) & 0.058 & 0.451 & 5.058 & 0.081 & 0.974 \\
EndoDAC\cite{Cui2024EndoDAC} & 0.052 & 0.370 & 4.582 & 0.074 & 0.976 \\
\textbf{This work} & \textbf{0.051} & \textbf{0.360} & \textbf{4.527} & \textbf{0.073} & \textbf{0.981} \\
\bottomrule
\end{tabular}

\end{table*}
\subsubsection{Comparison with State-of-the-Art}
\begin{figure*}[t]
    \centering
    \setlength{\tabcolsep}{1pt} 
    \renewcommand{\arraystretch}{0.5}
    \begin{tabular}{c c c c}
        & \small{AF-SfMLearner} & \small{EndoDAC} & \small{Ours} \\
        
        \rotatebox{90}{\small{Example 1}} & 
        \includegraphics[width=0.31\linewidth]{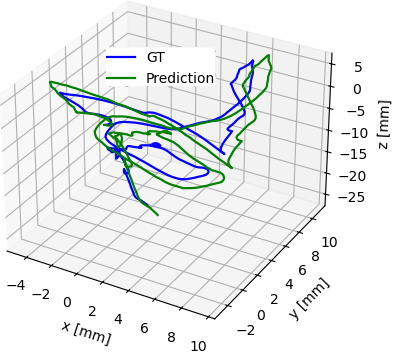} & %
        \includegraphics[width=0.31\linewidth]{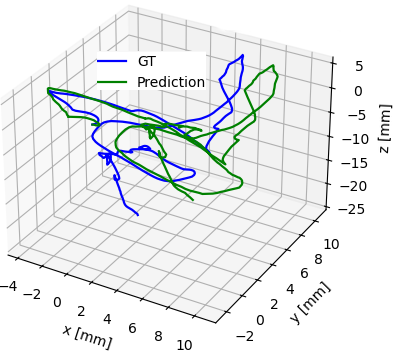} & %
        \includegraphics[width=0.31\linewidth]{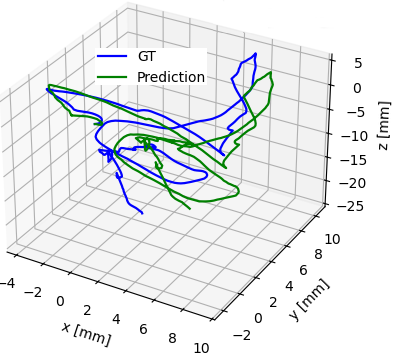} \\ %
        
        \rotatebox{90}{\small{Example 2}} & 
        \includegraphics[width=0.31\linewidth]{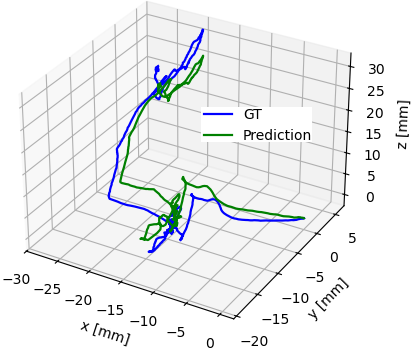} & %
        \includegraphics[width=0.31\linewidth]{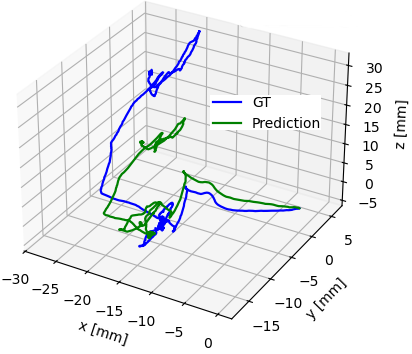} & %
        \includegraphics[width=0.31\linewidth]{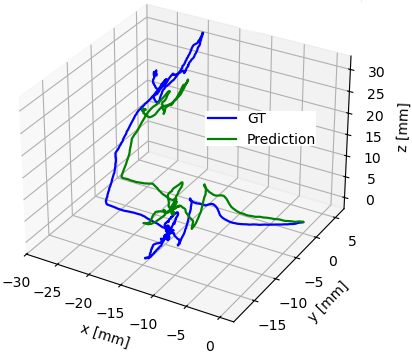} \\ %
        
    \end{tabular}
    
    \caption{Qualitative pose estimation comparison on the SCARED dataset. The rows correspond to Sequence 01 and Sequence 02. Our method (Right) significantly reduces trajectory drift compared to the baseline (Left), maintaining tighter alignment with the Ground Truth (Blue).}
    \label{fig:pose_qualitative}
\end{figure*}

Qualitative pose estimation comparisons on the SCARED dataset (Figure \ref{fig:pose_qualitative}) demonstrate that our method maintains tighter alignment with the ground truth compared to prior baselines. Further more, Table \ref{tab:sota_comparison} compares our method against established self-supervised baselines. On the aggregate test set, our framework achieves state-of-the-art accuracy, outperforming the strongest baseline \cite{Cui2024EndoDAC} and significantly surpassing traditional methods.

\begin{table*}[!ht]
\centering
\caption{Physically-stratified quantitative comparison. By clustering frames based on valid ground-truth density, we reveal that our method outperforms the EndoDAC baseline in the highly specular \textbf{Hard} and \textbf{Medium} categories, which represent the vast majority of the test set (475/551 frames).}

\begin{tabular}{llccccc}
\toprule
\textbf{Dataset} & 
\textbf{Method} & 
\textbf{Abs Rel} $\downarrow$ & 
\textbf{Sq Rel} $\downarrow$ & 
\textbf{RMSE} $\downarrow$ & 
\textbf{RMSE$_{\text{log}}$} $\downarrow$ & 
\textbf{$\delta < 1.25$ $\uparrow$} \\
\midrule

\multirow{2}{*}{\shortstack[l]{\textbf{Easy}\\$N=76$}} 
 & EndoDAC & \textbf{0.058} & \textbf{0.400} & \textbf{5.236} & \textbf{0.078} & 0.977 \\
 & \textbf{Ours} & 0.059 & 0.424 & 5.473 & 0.079 & \textbf{0.982} \\
\midrule
\multirow{2}{*}{\shortstack[l]{\textbf{Medium}\\$N=174$}} 
 & EndoDAC & \textbf{0.058} & 0.406 & 4.623 & 0.081 & 0.969 \\
 & \textbf{Ours} & 0.059 & \textbf{0.391} & \textbf{4.517} & \textbf{0.079} & \textbf{0.977} \\
\midrule
 \multirow{2}{*}{\shortstack[l]{\textbf{Hard}\\ $N=301$}} 
 & EndoDAC & 0.046 & 0.341 & 4.390 & 0.069 & 0.980 \\
 & \textbf{Ours} & \textbf{0.045} & \textbf{0.325} & \textbf{4.293} & \textbf{0.067} & \textbf{0.983} \\

\bottomrule
\end{tabular}

\label{tab:stratified_results}
\end{table*}

\begin{figure*}[!htbp]
    \centering
    \setlength{\tabcolsep}{1pt} 
    \renewcommand{\arraystretch}{0.5} 

    \begin{tabular}{m{1em}cccc}
        & \small Image & \small EndoDAC\cite{Cui2024EndoDAC} & \small This Work & \small Ground Truth \\[2pt]
        \rotatebox{90}{\small ~Easy~} & 
        \includegraphics[width=0.2\linewidth, valign=m]{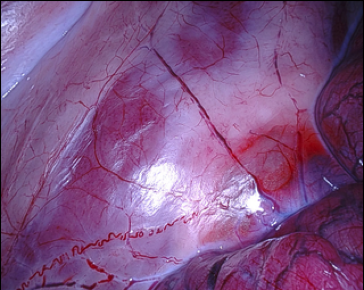} & 
        \includegraphics[width=0.2\linewidth, valign=m]{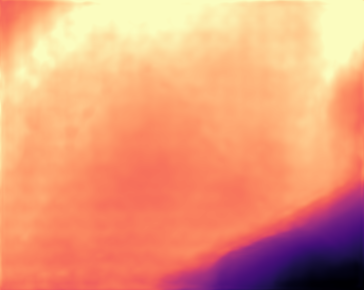} & 
        \includegraphics[width=0.2\linewidth, valign=m]{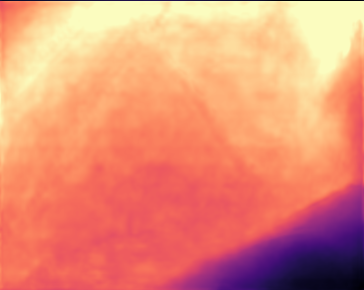} & 
        \includegraphics[width=0.2\linewidth, valign=m]{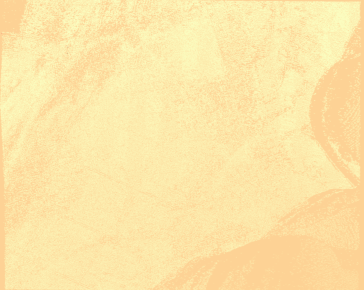} \\
        
        \rotatebox{90}{\small ~~Medium~~~} &
        \includegraphics[width=0.2\linewidth, valign=m]{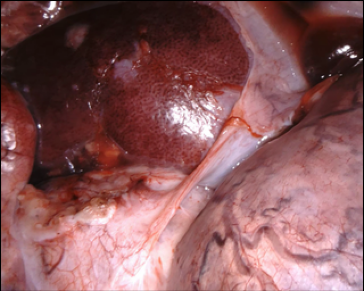} & 
        \includegraphics[width=0.2\linewidth, valign=m]{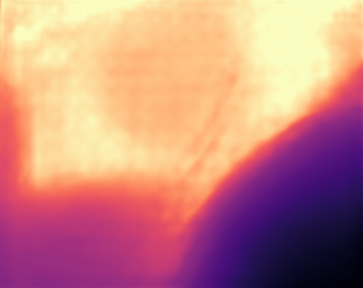} & 
        \includegraphics[width=0.2\linewidth, valign=m]{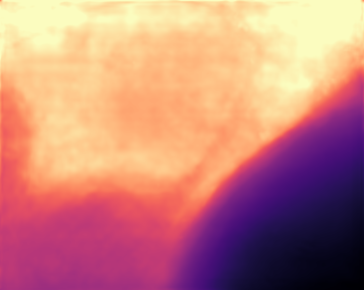} & 
        \includegraphics[width=0.2\linewidth, valign=m]{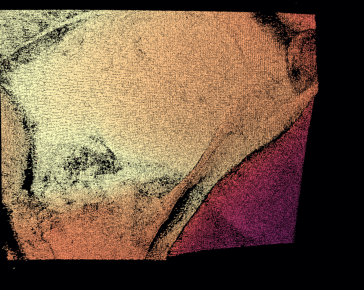} \\
        
        \rotatebox{90}{\small ~~~Hard~~~} &
        \includegraphics[width=0.2\linewidth, valign=m]{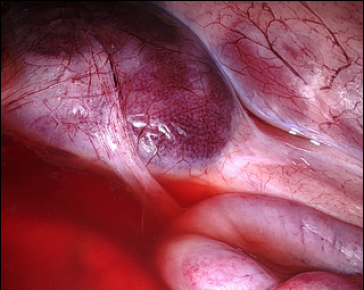} & 
        \includegraphics[width=0.2\linewidth, valign=m]{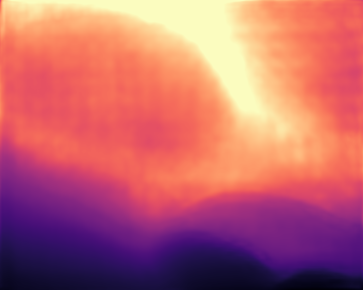} & 
        \includegraphics[width=0.2\linewidth, valign=m]{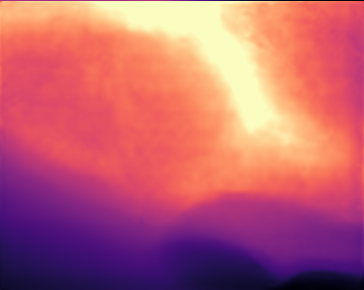} & 
        \includegraphics[width=0.2\linewidth, valign=m]{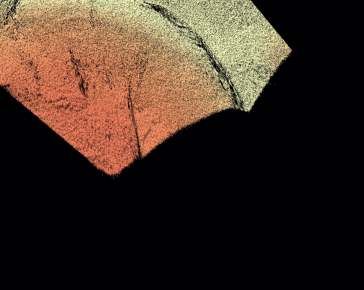} \\

    \end{tabular}
    \caption{Qualitative comparison across difficulty clusters (Physically-Stratified Protocol). The rows correspond to Easy (Top), Medium (Middle), and Hard (Bottom) subsets. Note how our method successfully preserves structural integrity and underlying geometry in the Hard (highly specular) regions, whereas the baseline model suffers from boundary collapse.}
    \label{fig:qualitative_h1}
\end{figure*}

\subsubsection{Robustness Analysis (Stratified Evaluation)}
The benefits of Synthetic Priors become evident when analyzing the physically-stratified results (Table \ref{tab:stratified_results}). In the \textbf{Hard (Specular)} cluster, our model outperforms EndoDAC \cite{Cui2024EndoDAC}, specifically reducing the Squared Relative Error from 0.341 to 0.325. This confirms that the synthetic pre-training allows the model to "see through" specular highlights that confuse real-world trained models. Figure \ref{fig:qualitative_h1} qualitatively demonstrates this advantage.

\begin{figure*}[!t] 
    \centering
    \setlength{\tabcolsep}{1pt} 
    \renewcommand{\arraystretch}{0.5} 

    \begin{tabular}{ccc}
        \small Image & \small EndoDAC\cite{Cui2024EndoDAC} & \small This Work  \\[2pt]
        \includegraphics[width=0.2\linewidth, valign=m]{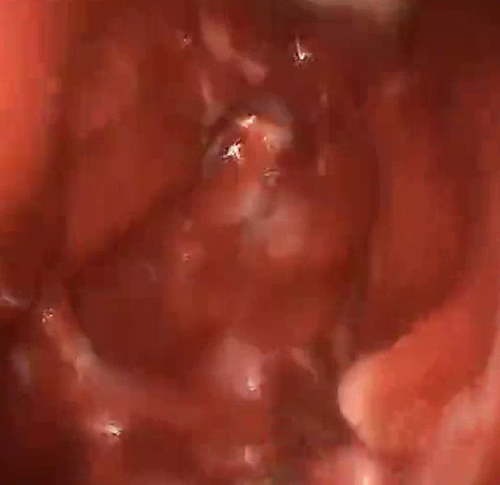} & 
        \includegraphics[width=0.32\linewidth, valign=m]{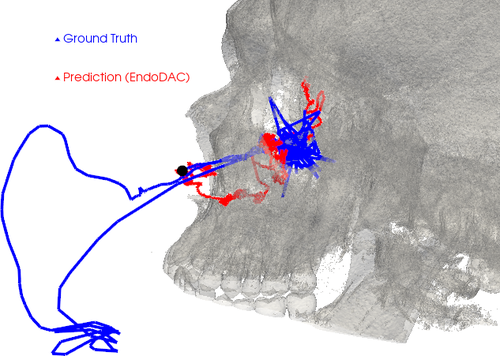} & 
        \includegraphics[width=0.32\linewidth, valign=m]{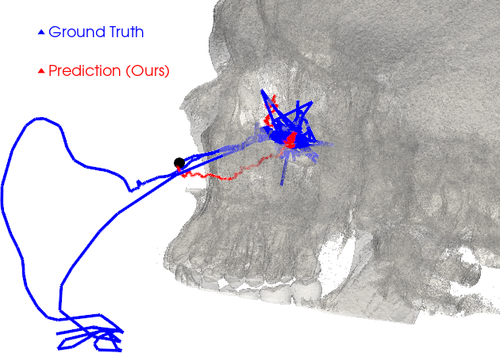} 

    \end{tabular}
    \caption{Comparison of 3D endoscopic trajectories on ROCAL-T 90. Ground truth (blue) sweeping deviations represent intentional out-of-body camera withdrawals, thus our evaluation focuses on the dense in-sinus operative cluster. Qualitatively, our algorithm generates a significantly more accurate and anatomically constrained trajectory (red) compared to the EndoDAC baseline, which suffers from severe drift and erroneously trespasses outside the surgical sites.}
    \label{fig:qualitative_trajectory}
\end{figure*}

\subsubsection{Performance on Baseline Failure Modes}
\begin{figure*}[!htbp]
        \centering
    \setlength{\tabcolsep}{1pt} 
    \renewcommand{\arraystretch}{0.5} 

    \begin{tabular}{m{1em}cccc}
        & \small Image & \small EndoDAC\cite{Cui2024EndoDAC} & \small This Work & \small Ground Truth \\[2pt]
        \rotatebox{90}{\small ~Easy~} & 
        \includegraphics[width=0.2\linewidth, valign=m]{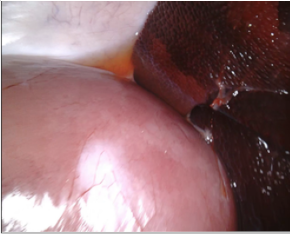} & 
        \includegraphics[width=0.2\linewidth, valign=m]{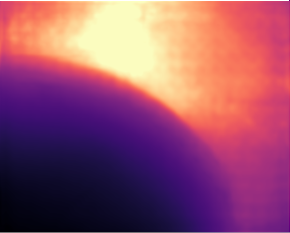} & 
        \includegraphics[width=0.2\linewidth, valign=m]{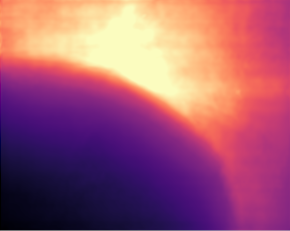} & 
        \includegraphics[width=0.2\linewidth, valign=m]{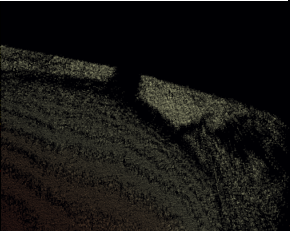} \\
        
        \rotatebox{90}{\small ~~Medium~~~} &
        \includegraphics[width=0.2\linewidth, valign=m]{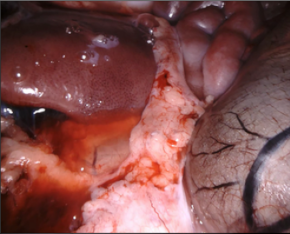} & 
        \includegraphics[width=0.2\linewidth, valign=m]{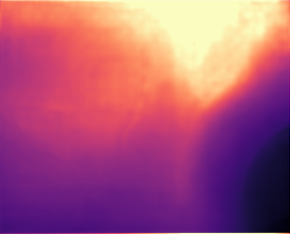} & 
        \includegraphics[width=0.2\linewidth, valign=m]{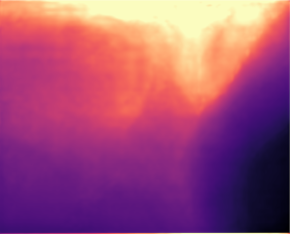} & 
        \includegraphics[width=0.2\linewidth, valign=m]{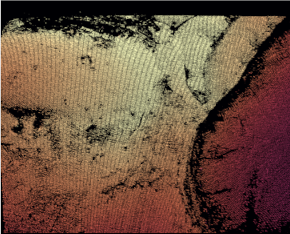} \\
        
        \rotatebox{90}{\small ~~~Hard~~~} &
        \includegraphics[width=0.2\linewidth, valign=m]{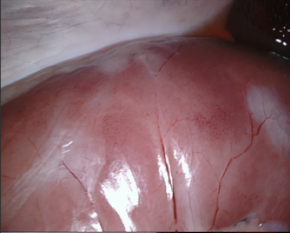} & 
        \includegraphics[width=0.2\linewidth, valign=m]{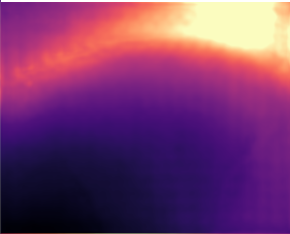} & 
        \includegraphics[width=0.2\linewidth, valign=m]{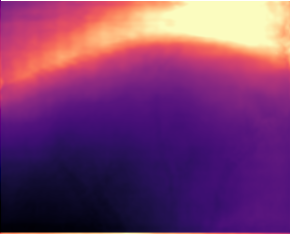} & 
        \includegraphics[width=0.2\linewidth, valign=m]{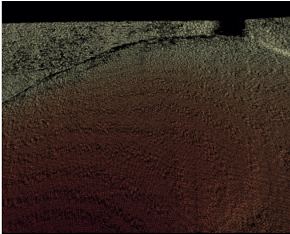} \\

    \end{tabular}
    \caption{Qualitative comparison of baseline failure modes. Frames are clustered by the error magnitude of the EndoDAC baseline, with rows corresponding to Easy (Top), Medium (Middle), and Hard (Bottom) subsets. Observe how our method effectively corrects the gross geometric distortions and boundary collapses present in the baseline's Hard subset.}
    \label{fig:qualitative_h2}
\end{figure*}

To further isolate the improvements, we stratified the test set based on the error distribution of the baseline model \cite{Cui2024EndoDAC} (Fig. \ref{fig:qualitative_h2}). As shown in Table \ref{tab:stratified_results_h2}, in frames where the V1 baseline fails most severely ("Hard", $\mu_{error} \approx 0.088$), our approach reduces the Squared Relative Error from 0.864 to 0.819 and RMSE from 7.152 to 6.893. This indicates that synthetic priors effectively correct the gross artifacts and boundary collapses inherent to real-world trained backbones.

\begin{table*}[!ht]
\centering
\caption{Performance on baseline failure modes. Test frames are clustered by the error magnitude of the EndoDAC baseline. Our method demonstrates significant recovery in the "Hard" cluster, indicating successful correction of the baseline's most severe failure cases.}
\label{tab:stratified_results_h2}
\begin{tabular}{llccccc}
\toprule
\textbf{Dataset} & 
\textbf{Model} & 
\textbf{Abs Rel} $\downarrow$& 
\textbf{Sq Rel} $\downarrow$& 
\textbf{RMSE} $\downarrow$& 
\textbf{RMSE$_{\text{log}}$} $\downarrow$& 
\textbf{$\delta < 1.25$ $\uparrow$} \\
\midrule
 \multirow{2}{*}{\shortstack[l]{\textbf{Easy}\\ $N=283$}} 
 & EndoDAC & \textbf{0.035} & \textbf{0.181} & \textbf{3.415} & \textbf{0.051} & 0.995 \\
 & \textbf{Ours} & 0.036 & 0.195 & 3.495 & \textbf{0.051} & 0.995 \\
\midrule
\multirow{2}{*}{\shortstack[l]{\textbf{Medium}\\ $N=197$}} 
 & EndoDAC & 0.060 & 0.462 & 5.328 & 0.088 & 0.970 \\
 & \textbf{Ours} & \textbf{0.059} & \textbf{0.431} & \textbf{5.156} & \textbf{0.084} & \textbf{0.978} \\
\midrule
 \multirow{2}{*}{\shortstack[l]{\textbf{Hard}\\$N=71$}} 
 & EndoDAC & 0.094 & 0.864 & 7.152 & 0.129 & 0.921 \\
 & \textbf{Ours} & \textbf{0.091} & \textbf{0.819} & \textbf{6.893} & \textbf{0.125} & \textbf{0.935} \\
\bottomrule
\end{tabular}

\end{table*}

\subsection{Clinical Validation on ROCAL-T 90}
To test the ultimate robustness of our synthetic priors, we evaluated the monocular pose tracking performance on the ROCAL-T 90 clinical dataset. 

\subsubsection{Rigorous Spatio-Temporal Alignment}
Evaluating monocular pose predictions against clinical hardware tracking requires a rigorous alignment pipeline. Because the hardware sensor and the endoscopic video operate asynchronously, we established the sensor's timestamps as the master clock. For every recorded hardware timestamp, the elapsed time was multiplied by the 24 fps video frame rate to isolate the exact synchronous video frame, extracting a precise 1-to-1 pairing of physical endoscope locations and network predictions. 

Furthermore, monocular pose networks suffer from inherent scale ambiguity. To resolve this for fair evaluation, both predicted and ground-truth trajectories were translated to a shared origin, and Singular Value Decomposition (SVD) was applied via the Kabsch algorithm to compute the optimal rotation matrix $R$. After globally scaling and aligning the predicted trajectory, we calculated the absolute 3D Euclidean distance (Absolute Trajectory Error, ATE) between the paired points at every synchronized time step $t$:
\begin{equation}
\resizebox{0.9\hsize}{!}{%
$\text{ATE}_t = \sqrt{(x_{gt,t} - x_{pred,t})^2 + (y_{gt,t} - y_{pred,t})^2 + (z_{gt,t} - z_{pred,t})^2}$%
}
\end{equation}

\subsubsection{Clinical Performance Analysis}
The quantitative results are presented in Table \ref{tab:ate_results}. Our DAv2-based model demonstrates quantifiable improvements over the baseline, reducing the ATE RMSE from 27.39 to 25.31 and the mean drift from 20.20 to 16.44. Qualitatively (Figure \ref{fig:qualitative_trajectory}), our method significantly reduces erratic trajectory drift compared to the baseline, maintaining a tighter correlation with the complex operative cluster.

\begin{table}[!htbp]
    \centering
    \caption{Quantitative evaluation of trajectory predictions. The Absolute Trajectory Error (ATE) is reported using both Root Mean Square Error (RMSE) and Mean drift. Lower values indicate better spatial alignment with the clinical ground truth.}
    \label{tab:ate_results}
    \begin{tabular}{lcc}
        \hline
        \textbf{Model} & \textbf{ATE RMSE} $\downarrow$ & \textbf{ATE Mean } $\downarrow$ \\
        \hline
        EndoDAC & 27.39 & 20.20 \\
        This work & 25.31 & 16.44 \\
        \hline
    \end{tabular}
\end{table}

However, a critical observation from plotting the full clinical trajectories against the ground truth is the substantial absolute deviation exhibited by both algorithms. While our synthetic priors significantly mitigate local structural failures, large sweeping deviations—often corresponding to external, out-of-body movements or rapid camera withdrawals typical of surgical workflows—remain highly erroneous. Rather than a mere failure, the absolute error actively exposes the ex-vivo to in-vivo domain shift problem. It highlights the extreme difficulty of purely monocular tracking in unconstrained clinical settings and establishes ROCAL-T 90 as a challenging, necessary benchmark to drive future iterations of surgical SLAM architectures.

\section{Conclusion}
In this work, we addressed the critical ex-vivo to in-vivo gap in endoscopic monocular depth estimation, where models trained on real-world pseudo-labels fail in specular, fluid-filled clinical environments due to severe label noise. To overcome this limitation, we proposed a paradigm shift toward leveraging the high-fidelity synthetic priors of the \textit{Depth Anything V2} architecture. By efficiently integrating these priors using Dynamic Vector LoRA, we successfully transferred precise, noise-free geometric representations to the surgical domain with a minimal parameter budget. 

Furthermore, to rigorously quantify model robustness, we introduced two distinct evaluation frameworks. First, we proposed a physically-stratified evaluation protocol on the public SCARED dataset, demonstrating that our approach significantly outperforms established baselines in high-specularity regimes. Second, and most crucially, we introduced ROCAL-T 90, the first clinical validation benchmark providing continuous in-vivo endoscopic sequences aligned with preoperative CT for sub-millimeter ground-truth trajectories. While our synthetic-prior framework establishes a new state-of-the-art and reduces trajectory drift in these true operative settings, the residual absolute errors actively expose the extreme difficulty of unconstrained clinical SLAM. Ultimately, by releasing ROCAL-T 90, we not only validate our approach but also provide the robotics community with a necessary and highly challenging benchmark to drive the next generation of clinically viable surgical perception systems.

\section*{Acknowledgments}
This research was partially funded by the NIH under grant number 1R15EB034519-01A1 and  NSF under grant number 2346790.
The data collection and usage of this work have been approved by University of Rochester IRB STUDY9047. 
\balance
\bibliographystyle{ieeetr}
\bibliography{\bibpath/robot,\bibpath/machinelearning,\bibpath/mine,\bibpath/causal,\bibpath/vision,\bibpath/3drecon,references} 

\end{document}